\documentclass{article}
\usepackage[utf8]{inputenc}
\usepackage{amssymb}
\usepackage{amsmath}
\usepackage{wasysym}
\usepackage{float}
\usepackage{hyperref}
\usepackage{wrapfig}
\usepackage{graphicx}
\usepackage{mwe}
\usepackage{geometry}
\title{WALL-E Technical Report}
\author{Tianbing Xu (Baidu Research), Andrew Zhang (Stanford), Liang Zhao (Baidu Research)}
\date{September 2018}

\begin{document}

\maketitle

\section{Motivation}
We release our codebase  (\href{WALL-E}{https://github.com/tianbingsz/WALL-E}), 
an efficient, fast, yet simple Reinforcement Learning Research Framework with potential applications in robotics and beyond.
There are two halves to RL systems: experience collection time and policy learning time. For a large number of samples in rollouts, experience collection time is the major bottleneck. Thus, it is necessary to speed up the rollout generation time with multi-process architecture support.

\section{Introduction}
\begin{figure}[h]
    \centering{\includegraphics[scale=0.35]{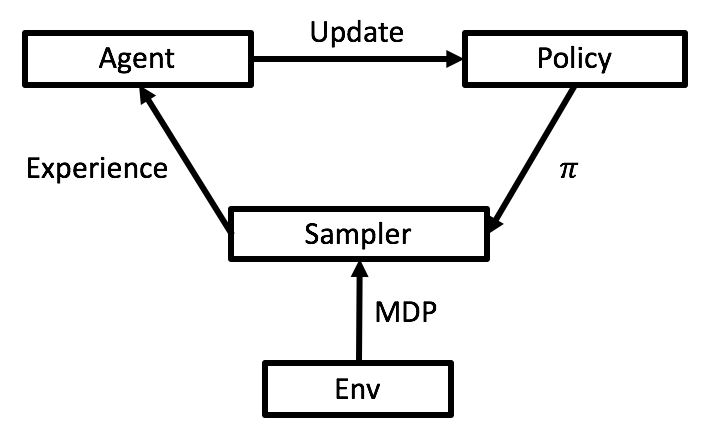}}
    \caption{RL flow chart}
    \label{fig:rl}
\end{figure}

Overall, reinforcement learning (RL) involves an agent interacting with an environment through repeatedly running a policy $\pi$, collecting experience from each iteration and using that experience to update its policy for maximal reward (Fig \ref{fig:rl}).
\\ Thanks to advancements in big data, computing power, and other machine learning discipline, reinforcement learning has emerged as the pinnacle field in pushing humanity closer to true artificial intelligence. Model-based reinforcement learning, for example, aims to build an accurate model (such as a MDP) of the environment dynamics and train the agent on said model, giving model learning capabilities as well as ease of reward learning. On the other hand, in model-free reinforcement learning, the agent does not have explicit information regarding state transitions and must continuously explore and generate experience to find the optimal policy.\\\\
In recent years, major problems have arisen in the field of reinforcement learning, such as planning and how to balance exploration and exploitation. Of particular interest, however, is the problem of knowledge gathering, namely how to efficiently and quickly sample trajectories to gain experience and update the policy without adversely affecting average return.

\section{Architecture}
\begin{figure}[t]
    \centering{\includegraphics[scale=0.35]{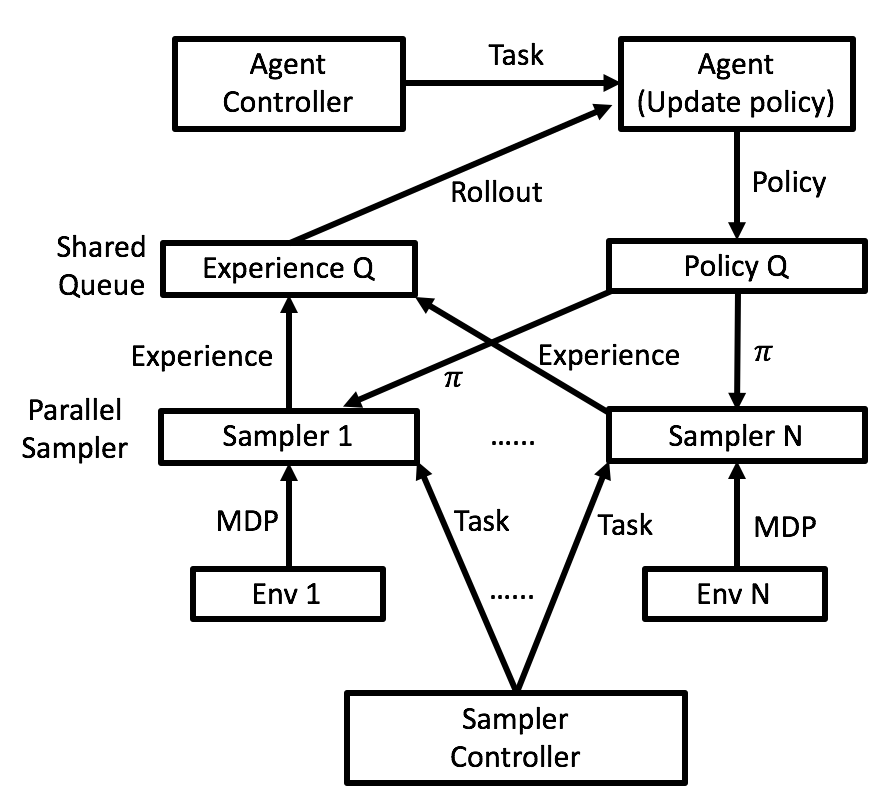}}
    \caption{Parallel Sampler RL Architecture}
    \label{fig:parallel}
\end{figure}

Our work, dubbed WALL-E, utilizes multiple rollout samplers running in parallel to rapidly generate experience. For starters, the agent processor runs asynchronously and updates the policy based on experience from the experience queue when ready, sending policy parameters to the policy queue. In turn, the $N$ sampler processors, concurrently generate experience based on the updated policy read from the primed policy queue and sends experience back to the experience queue (Fig \ref{fig:parallel}).\\
\\\\\\\\\\
\section{Results}
\begin{figure}[h]
    \centering{\includegraphics[scale=0.35]{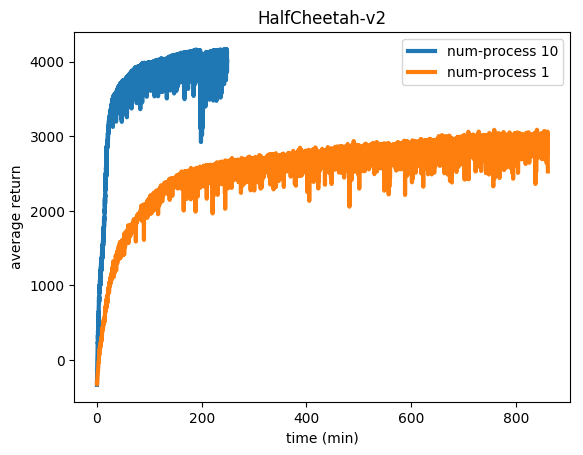}}
    \caption{Comparison of $N=10$ vs $N=1$}
    \label{fig:rewardresults}
\end{figure}

\begin{figure}[h]
    \centering{\includegraphics[scale=0.35]{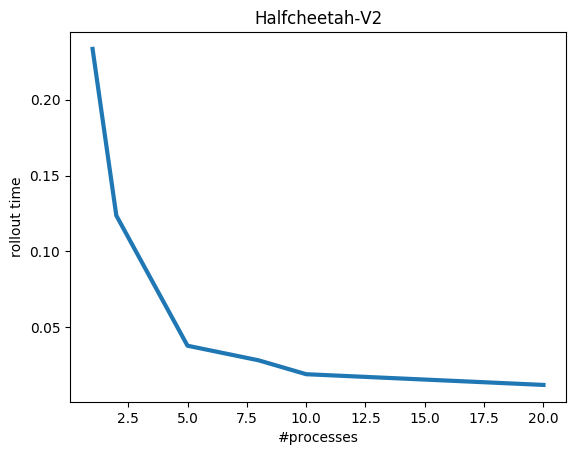}}
    \caption{Rollout time decrease}
    \label{fig:rollouttimes}
\end{figure}

\begin{figure}[t]
    \centering{\includegraphics[scale=0.35]{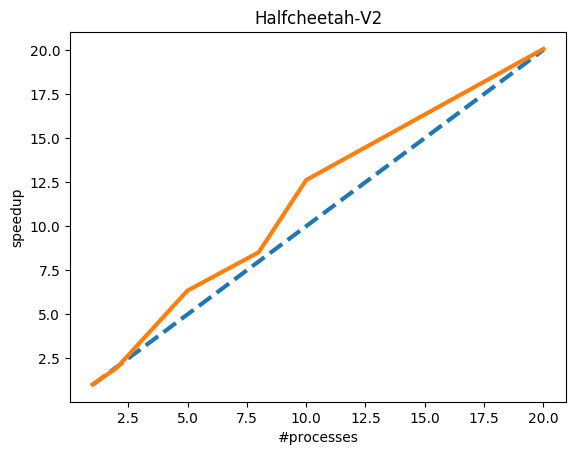}}
    \caption{Speedup comparison}
    \label{fig:speedup}
\end{figure}

\begin{figure}[h]
    \centering
    \begin{minipage}{.5\textwidth}
        \centering{\includegraphics[scale=0.35]{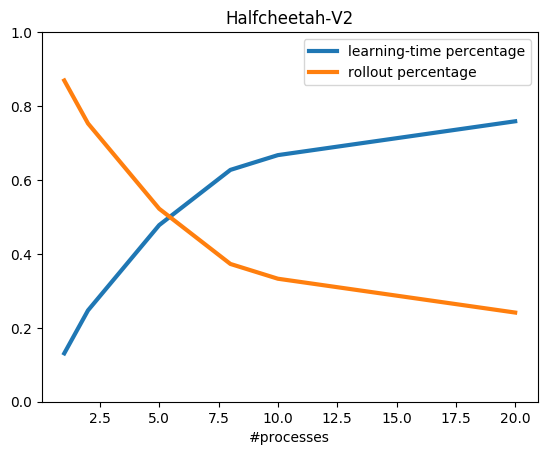}}
        \caption{Percentage of Policy Learning and Experience Collection time w.r.t. num CPUs}
        \label{fig:percentage}
    \end{minipage}\hfill
    \begin{minipage}{.5\textwidth}
        \centering{\includegraphics[scale=0.35]{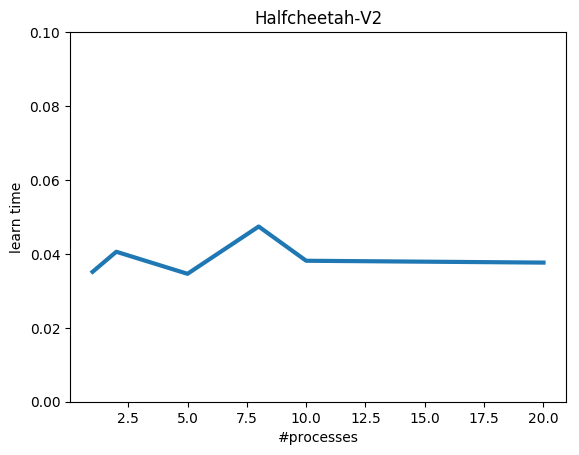}}
        \caption{Policy Learning time for each iteration w.r.t num CPUs}
        \label{fig:policylt}
    \end{minipage}
\end{figure}

Due to our parallel samplers, we experience not only faster convergence times, but also higher average reward thresholds. For example, on the MuJoCo HalfCheetah-v2 task, with $N = 10$ parallel sampler processes, we are able to achieve much higher average return than those from using only a single process architecture (Fig \ref{fig:rewardresults}). 
\\ By testing other values of $N$ on $20000$ samples per iteration, we notice a significant decrease in rollout time w.r.t. processor count (Fig \ref{fig:rollouttimes}) that surprisingly does not negatively impact the average return in a non-trivial fashion.
\\ Furthermore, by plotting the speedup from $20000$ samples per iteration (Fig \ref{fig:speedup}), we see that the running time is highly fast with a variance from the asynchronous nature and the queue I/O. Thus, based on our results, we conclude that the experience collection speedup w.r.t. to CPU numbers is near-linear (while not over-linear).
\\\\\\
As we can see from Fig.\ref{fig:percentage}, with the near-linear decrease of experience collection time w.r.t. the number of CPUs, the data collection is no longer the bottleneck while the percentage of policy learning time increases to become the next bottleneck, though the overall policy learning time is almost keeping the same for each iteration from Figure \ref{fig:policylt}.

\section{Related Work}
We would like to thank Danijar Hafner, James Davidson, and Vincent Vanhoucke [2]
for their excellent work that inspired these ideas about speeding up experience collection for RL.
Kevin Frans and Danijar Hafner [1]
 introduced a similar parallel architecture focused on speeding up experience collection. Instead of using proximal policy optimization algorithms, they utilized the trust-region policy optimization algorithm, whereby each actor process contained a copy of the policy as well as the environment and returned transition experience for each timestep. Due to the asynchronous nature of their system, they forego the step of running episodes until a certain number of timesteps in the original TRPO algorithm and alternatively estimate the number of episodes by dividing a desired number of timesteps by the average episode length using the previous policy. 
Todd Hester, Michael Quinlan and Peter Stone also developed a multi-thread architecture [3, 4] 
for model-based RL that runs in real time by parallelizing model learning, planning and acting.
 
\section{Further Work}
This is only the starting point of our long-term research project.  The next step is to make the codebase generalize enough to support important RL algorithms (On-Policy, Off-Policy and so on) and our on-going RL research.
For example, but not restrict to,
\begin{enumerate}
    \item Off-policy learning (DDPG) with replay buffer, as Off-Policy learning requires much more samples than policy gradient methods, it might be an advantage to adopt the parallel experience collection architecture.
    \item Parallel computation of policy learning, since we have reduced the experience collection time, the policy learning becomes the relative bottleneck and it is worth trying to distribute the policy update on different CPUs or GPUs.
\end{enumerate}


\end{document}